\documentclass[preprint,12pt]{elsarticle}

\usepackage{amsmath,amssymb,amsfonts}
\usepackage[linesnumbered,algoruled,boxed,lined]{algorithm2e}
\usepackage{subfigure}
\usepackage{setspace}
\usepackage{booktabs}
\usepackage{multirow}
\usepackage{threeparttable}
\usepackage{array}
\usepackage{algorithmic}
\usepackage{graphicx}
\usepackage{textcomp}
\usepackage{multirow}
\usepackage{float}
\usepackage{arydshln}
\usepackage{mathrsfs}
\usepackage{verbatim}

\usepackage{lineno,hyperref}

\usepackage{graphics}
\usepackage{epsfig}

\modulolinenumbers[5]

\journal{}

\begin{document}

\begin{frontmatter}

\title{Semi-Supervised Crowd Counting from Unlabeled Data}

\author[Newcastle_Computing]{Haoran~Duan}\ead{michael.duan28@gmail.com}
\author[Durham]{Fan~Wan\#}\ead{Fan.Wan@durham.ac.uk}
\author[Newcastle_Computing]{Rui~Sun\#}\ead{r.sun5@newcastle.ac.uk}
\author[dalian]{Zeyu~Wang\corref{mycorrespondingauthor}}\ead{20231578@dlnu.edu.cn}
\author[Newcastle_Computing]{Varun~Ojha}\ead{varun.ojha@newcastle.ac.uk}
\author[Newcastle_Computing]{Yu Guan}\ead{yu.guan@newcastle.ac.uk}
\author[Durham]{Hubert~P. H. Shum}\ead{hubert.shum@durham.ac.uk}
% \author[CAS1]{Shizheng Wang}\ead{shizheng.wang@foxmail.com}
\author[CAS2]{Bingzhang Hu}\ead{hubzh@aiofm.ac.cn}
% \author[Tencent]{Yawen Huang}\ead{yawenhuang@tencent.com}
\author[Durham]{Yang Long\corref{mycorrespondingauthor}}\ead{yang.long@durham.ac.uk}
% \author[Newcastle_Computing]{Rajiv~Ranjan}\ead{raj.ranjan@newcastle.ac.uk}
% \author[Tencent]{Yefeng Zheng}\ead{yefengzheng@tencent.com}
% \address{Radarweg 29, Amsterdam}
% \fntext[myfootnote]{Since 1880.}

% \author[mymainaddress,mysecondaryaddress]{Elsevier Inc}

% \author[mysecondaryaddress]{Global Customer Service\corref{mycorrespondingauthor}}
\address[Durham]{Department of Computer Science, Durham University, UK.}
\address[dalian]{College of Computer Science and Engineering, Dalian Minzu University.}
% \address[Newcastle_Engineering]{School of Engineering, Newcastle University, UK.}
\address[Newcastle_Computing]{School of Computing, Newcastle University, UK.}
% \address[Tencent]{Jarvis Research Center, Tencent YouTu Lab, China.}
\address[CAS2]{Institute of Physical Science, Chinese Academy of Science, Hefei, China}
% \cortext[equal contribution]{Co-First Author,$^{\#}$ equal contribution}
\cortext[mycorrespondingauthor]{Corresponding author $^{\#}$ equal contribution}

\begin{abstract}
Automatic Crowd behavior analysis can be applied to effectively help the daily transportation statistics and planning, which helps the smart city construction. As one of the most important keys, crowd counting has drawn increasing attention. Recent works achieved promising performance but relied on the supervised paradigm with expensive crowd annotations. To alleviate the annotation cost in real-world transportation scenarios, in this work we proposed a semi-supervised learning framework $S^{4}\textit{Crowd}$, which can leverage both unlabeled/labeled data for robust crowd counting. In the unsupervised pathway, two \textit{self-supervised losses} were proposed to simulate the crowd variations such as scale, illumination, based on which supervised information pseudo labels were generated and gradually refined.
We also proposed a crowd-driven recurrent unit \textit{Gated-Crowd-Recurrent-Unit (GCRU)}, which can preserve discriminant crowd information by extracting second-order statistics, yielding pseudo labels with improved quality. 
A joint loss including both unsupervised/supervised information was proposed, and a dynamic weighting strategy was employed to balance the importance of the unsupervised loss and supervised loss at different training stages. 
We conducted extensive experiments on four popular crowd counting datasets in semi-supervised settings. 
Experimental results supported the effectiveness of each proposed component in our $S^{4}$Crowd framework. 
Our method achieved competitive performance in semi-supervised learning approaches on these crowd counting datasets.
\end{abstract}

\begin{keyword}
Computer Vision, Crowd Counting, Semi-Supervised Learning, Self-Supervised Learning, Deep Learning.
\end{keyword}

\end{frontmatter}

% \linenumbers

\section{Introduction}
Crowd counting plays a vital role in real-world applications, specifically it has had obviously increasing impact for intelligent transportation in smart cities \cite{wang2022crowd,liu2020nssnet,ding2020crowd,zaki2017automated}. Automatic crowd counting can directly help the human operators to dynamically and accurately obtain the real-time statistics of objects (i.e., humans trasffic) traffic or the level of congestion in the various conditions of transportation. With the help of the crowd counting, transportation planning can be effectively adjusted and some traffic problems can be quickly solved or avoided.

In this work, we focus on image-based crowd scenes (captured from public surveillance) estimation with explicit pedestrians counting. The ways of obtaining the crowd statistics have been progressively explored from detection-based \cite{viola2005detecting}, segmentation-based approaches \cite{zhao2008segmentation} to head density estimation based methods \cite{lempitsky2010learning}. 
Compared with detection/segmentation-based approaches, density-based methods tended to be less sensitive to occlusions in dense crowd, and they became the mainstream approaches nowadays.

For density-based methods, Convolution Neural Network (CNN) was the major technique \cite{hu2020count,jiang2020attention} for crowd representation learning and density estimation.
While existing CNN-density-based crowd counting methods achieved promising performance, they relied heavily on labeled data, and the acquisition of labels can be labour-intensive and time-consuming. 
For example, it took 2000 human-hours to annotate 1535 images (with 1251642 locations of crowd heads) in the UCF\_QNRF dataset \cite{idrees2018composition}.
When with inadequate annotated labels, overfitting may occur.

To reduce the overfitting effect and the annotation costs \cite{wang2022crowd,liu2020nssnet,ding2020crowd,zaki2017automated}, semi-supervised learning, which can take advantage of both labelled/unlabelled data, has been widely used in many computer vision tasks such as semantic segmentation \cite{mittal2019semi}, object detection \cite{zhao2020sess}, action recognition \cite{si2020adversarial}.
Most recently, semi-supervised learning was also employed in the crowd counting community.
Sindagi et al. proposed a Gaussian-Process iterative learning to estimate the pseudo labels for unlabeled crowd data \cite{sindagi2020learning}, and Liu et al. proposed to learn the crowd representation from unlabeled data with some surrogate segmentation tasks \cite{liu2020semi}. 
Both works\cite{sindagi2020learning, liu2020semi} demonstrated the effectiveness of leveraging the unlabelled crowd data, yet their semi-supervised schemes failed in modelling the variations of crowds (e.g., head scale, illumination, perspective distortion, etc.), which are considered as indispensable factors for reliable crowd modelling.

To address this issue, we proposed to model different crowd variations in a self-supervised manner, specifically with two regularization terms. In the meantime, the pseudo labels of the un-annotated data were generated and gradually refined for better model generalization. In our semi-supervised scheme, a joint cost function including both supervised/unsupervised loss was proposed, and a dynamic weighting strategy was employed to balance both losses at different training stages.

\begin{figure}[th]
\centering
\includegraphics[width=0.95\linewidth]{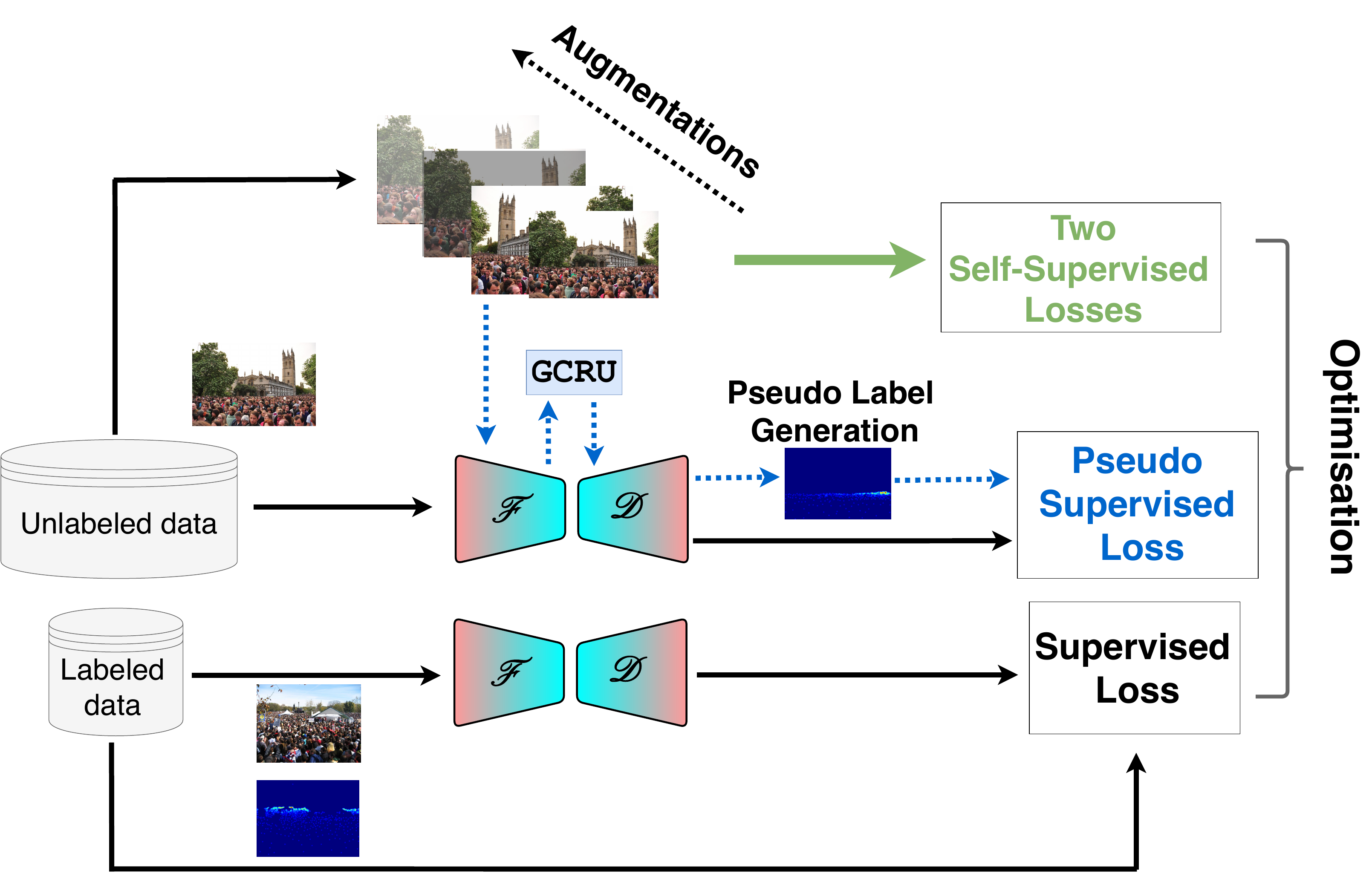}
\caption{The Proposed $S^{4}$Crowd Framework. It aims to provide reliable pseudo labels for unlabeled data utilizing series output, which is used to train the model along with the labeled data. 
}
\label{fig:architecture}
\end{figure}
Specifically in the unsupervised pathway, a sequence of augmentation functions (i.e., image transformations) were applied to each unlabeled image for simulating different variations.
To better encode the high-order-tensor augmented image sequences for high-quality pseudo labels generation, we proposed a novel recurrent unit, namely Gated-Crowd-Recurrent-Unit (GCRU). Compared with the conventional GRU\cite{cho2014learning}, GCRU further considered the second-order statistics of crowd, which was demonstrated as informative features in crowd modelling \cite{duan2020sofa}.
The general structure of our proposed $S^{4}$Crowd is shown in Fig.\ref{fig:architecture}.
In this work, our main contribution can be summarised as:

\begin{itemize}
\item We proposed a novel semi-supervised crowd counting framework  $S^{4}$Crowd, which included several key components (variations modelling, pseudo labeling, dynamic weight training scheme.) for robust crowd modelling when with limited labeled data. 
\item We proposed two self-supervised losses/regularisation terms, i.e., Crowd Scale Equivariance (CSE), Crowd Entropy Consistency (CEC), based on which we modelled the crowd variations in an unsupervised manner. 
\item A crowd-driven recurrent unit, i.e., Gated-Crowd-Recurrent-Unit (GCRU) was proposed to encode high-order-tensor crowd sequences, which can also learn the informative second-order crowd statistics for better crowd modelling. 
\item Extensive experiments were conducted, and the proposed components in our semi-supervised $S^{4}$Crowd framework were well studied. Our method outperformed other semi-supervised crowd counting algorithms in various settings.  
\end{itemize}

\section{Related Work}
% \subsection{Crowd Counting} 
Lempitsky and Zisserman casted the crowd counting problem as the efficient density estimation of only head area \cite{lempitsky2010learning}, and the integrating over density map gives the counting of crowd. Furthermore, CNN-based methods have been one of the main techniques for density-based crowd counting \cite{Cheng_2019_ICCV,hu2020count, zeng2021multi} to learn the crowd representation and estimate the density maps. Those methods normally adopt the following strategies, such as multi-scale information fusion\cite{cao2018scale,jiang2019crowd} that incorporates contextual semantic information and attention-guided approaches\cite{zhang2019relational,zhang2019attentional,duan2023dynamic} that use attention mechanisms to focus on relevant information. Additionally, multi-task learning-based approaches\cite{jiang2020attention,luo2020hybrid,sunsora,wan2024sentinel} that leverage auxiliary tasks like foreground and background segmentation\cite{zhao2019leveraging}, depth predictions\cite{lian2019density}, and uncertainty estimation\cite{oh2020crowd} have been developed to further enhance model performance. However, the above supervised methods require a substantial quantity of annotated data, which is a time-consuming and labor-intensive process. To alleviate additional annotation cost for crowd model training, semi-supervised crowd counting \cite{liu2020semi,sindagi2020learning, zhao2020active} has drawn some attentions. Specifically, a Gaussian-Process (GP) was used to model the relationship between the feature latent space and ground truth \cite{sindagi2020learning} in labeled data, and generated pseudo labels for unlabeled data. In \cite{liu2020semi}, Liu et al. proposed a self-training framework with surrogate segmentation tasks, and the binary inter-relationship of crowd and non-crowd leads to the accurate density prediction. Also, Zhao et al. proposed a active learning framework \cite{zhao2020active} to select a small fraction of most informative data to be annotated to train the model.

Semi-supervised learning has been an active area of research in recent years, with several new methods being proposed for effective utilization of unlabeled data. One such method is Triple Generative Adversarial Networks (Triple-GAN) \cite{zhang2018triple}, which leverages the power of adversarial training to improve the quality of generated samples and achieve state-of-the-art results on several benchmark datasets. Another popular approach is MixMatch \cite{berthelot2019mixmatch}, which uses a combination of supervised and unsupervised losses to achieve improved performance on low-data regimes. In the medical domain, several semi-supervised learning methods have been proposed, such as Multi-Task Self-Supervised Learning (MTSSL) \cite{wang2020multi}, and DivideMix \cite{li2020dividemix}, which achieve state-of-the-art results on medical imaging datasets. Robust semi-supervised learning methods have also been proposed to handle outliers and noisy data, such as Robust PCA \cite{guan2012robust} and Zero-Shot Learning \cite{long2017zero}.
In addition to data augmentation, another popular technique for semi-supervised learning is Pseudo-Labeling \cite{Lee2013PseudoLabelT,xie2020self}, which involves assigning labels to unlabeled data points based on the predictions of a trained model. This approach has been shown to be effective in several settings and has been extended to handle noisy labels and other challenges. Overall, recent work in semi-supervised learning has focused on improving the quality of generated samples, handling noisy and outlier data, and developing new regularization techniques to improve consistency and generalization. These advances have the potential to significantly improve the performance of intelligent transportation systems, as well as other applications where large amounts of unlabeled data are available.
% Every day even every second, there is a massive volume of data streamed from traffic monitoring or public surveillance and it may not be possible to annotate it. Therefore, successfully applying semi-supervised learning is valuable, which can improve the capability of intelligent transportation systems via not only existing limited labelled data but also the lots of unlabelled data.

\section{Methodology}

\begin{figure*}
\centering
\includegraphics[width=\linewidth]{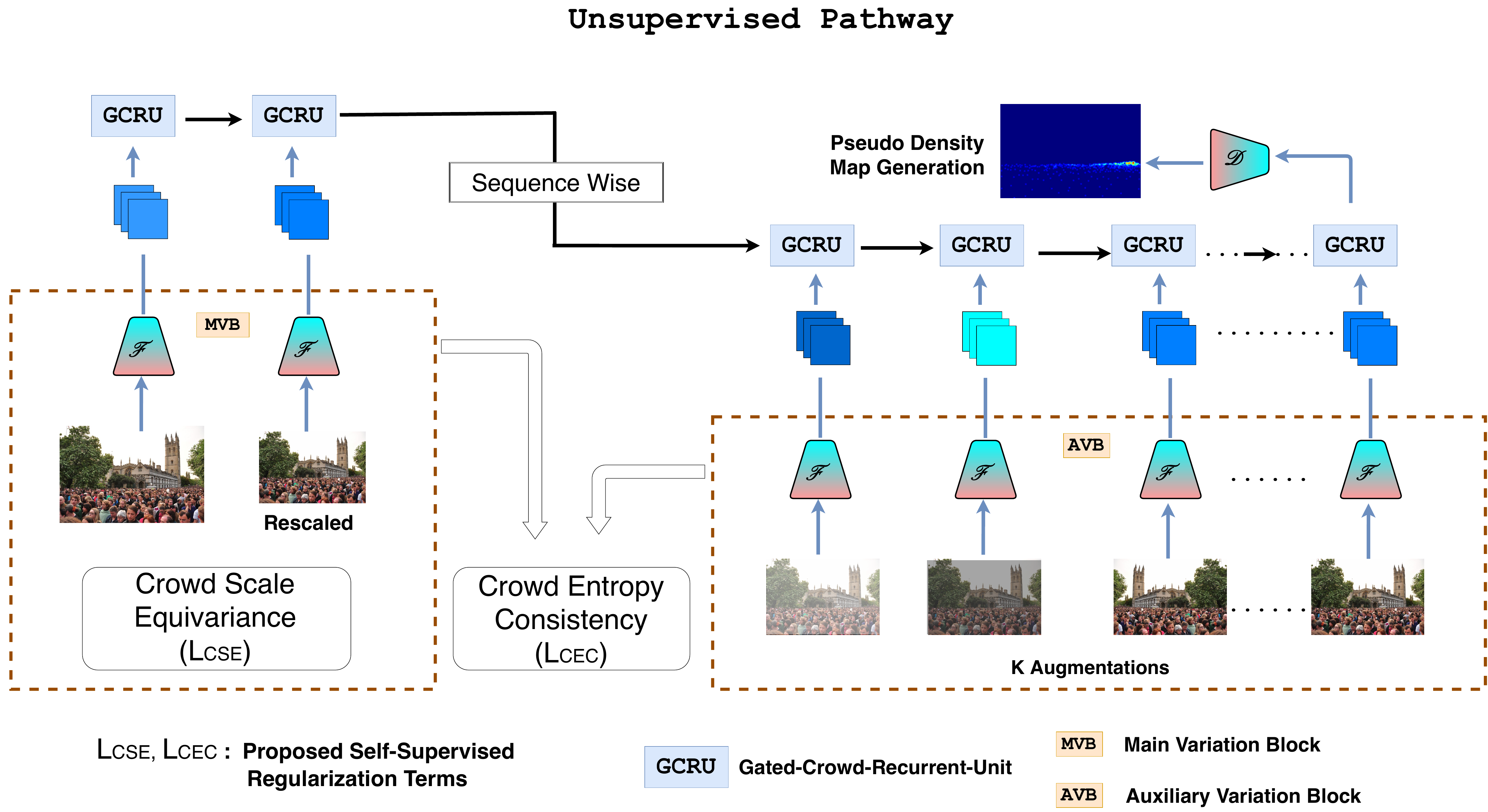}
\caption{The overall architecture of the proposed Unsupervised pathway, which consists of CSE/CEC regularization terms and pseudo labels generation with Gated-Crowd-Recurrent-Unit(GCRU).}
\label{fig:unsup-path}
\end{figure*}

\subsection{Problem Statement}

%Recently, Supervised 
Density-based crowd counting has been regarded as a regression problem \cite{hu2020count,lempitsky2010learning}, and pixel-wise density should be estimated before counting.
Given a crowd image sample $\mathbf{x}$ with $e$ pedestrians' heads $\mathbf{V} = \{ \mathbf{v}_{q} \in \mathbb{N}^{2} \}^{e}_{q=1}$, where $\mathbf{v}_q$ is the x-y coordinate of the ${q}_{th}$ head's center point, and the ground truth density map of $\mathbf{x}$ can be constructed by $e$ Gaussian distributions ($\mathcal{N}$) with $\mathbf{V}$ as the mean values.  %over all the head pixels in image $\mathbf{x}$, 
Then the crowd can be counted by calculating the
%the crowd counts can then be calculated by 
the integral of the density map. The ground truth density map $\mathbf{y}^{gt}$ can be formed as:

\begin{equation}
\mathbf{y}^{gt} =\sum_{\mathbf{v} \in \mathbf{V}} \mathcal{N}\left(p ; \mu=\mathbf{v}, \sigma^{2}\right)
\end{equation}

\noindent{where $ p \in \mathbf{x}$ denotes the image pixels, $\sigma$ is a fixed \cite{jiang2020attention} value with a very small number spanning a few pixels \cite{lempitsky2010learning}, and the head counting can be calculated via $e = \sum_{p \in \mathbf{x}} \mathbf{y}^{gt}_{p}$}. 

A typical crowd density estimation model normally included a 
%During the training, the crowd counting model contains a 
crowd representation learning module $\mathscr{F}$ and a density regression module $\mathscr{D}$ \cite{li2018csrnet,jiang2020attention}. 
Given the predicted density map $\mathbf{y}^{pr} = \mathscr{D}(\mathscr{F}(\mathbf{x}))$ , the supervised training process is to minimise the following MSE loss \footnote[1]{Note for demonstration, we only used sample-wise loss here}
%the model is supervised by a pixel-wise loss $\mathcal{L}_{sup}$, which is formed as:
\begin{equation}
\mathcal{L}_{S} = \frac{1}{WH} \sum_{w=1}^{W} \sum_{h=1}^{H}\left(\mathbf{y}^{pr}_{w,h}-\mathbf{y}^{gt}_{w,h}\right)^{2}
\label{eq:L_s}
\end{equation}

%with height ($H$) and width ($W$)
\noindent{where $H$ and $W$ are image height and width respectively. In the inference stage, given the trained model and a query crowd image $\mathbf{x}'$, the predicted density map $\mathbf{y}^{pr}$ and the corresponding counts $e^{pr}$ can be calculated as follows:}

\begin{equation}
e^{pr} = \sum_{p \in \mathbf{x}'} \mathbf{y}^{pr}_{p}, \text{where} \ \mathbf{y}^{pr} = \mathscr{D}(\mathscr{F}(\mathbf{x}')).
\label{eq:inference}
\end{equation}

\noindent{However, the aforementioned supervised pathway may require costly annotations. 
When with inadequate labelled data, overfitting may occur, yielding unreliable $\mathscr{F}$ and $\mathscr{D}$ for crowd modelling.  
%fully supervised crowd counting requires costly annotation, and learning from limited labeled data is hard to reduce the effect of crowd variations. 

There are several ways to alleviate the overfitting effect without additional annotation requirement, such as regularisation, unsupervised learning (e.g., self-supervised learning), semi-supervised learning, etc. 
%There were several ways to alleviate costly annotation and improve model's generalization (e.g. transfer learning, self-supervised learning, semi-supervised learning, etc.). 
%In this work, we proposed a semi-supervised framework, which can take advantage of both unlabelled/labelled data for 
For reliable $\mathscr{F}$ and $\mathscr{D}$ estimation, in this work we proposed a semi-supervised framework, where both unsupervised and supervised losses were combined for model training.  
%we proposed a Pseudo-Labeling-based \cite{Lee2013PseudoLabelT,xie2020self,berthelot2019mixmatch} semi-supervised framework. %with self-supervised regularization, which can take advantage of both unlabelled/labelled data.
%After supervised initialisation, 
%We used both supervised and unsupervised learning to generate pseudo labels. {\color{blue} precise?} 
Specifically, in the unsupervised learning pathway two self-supervised losses were derived to model the crowd variations, based on which a crowd-driven recurrent unit GCRU was proposed to better encode crowd sequential information for pseudo label generation. 
Based on the pseudo labels, the trained  
$\mathscr{F}$ and $\mathscr{D}$ can be more robust to crowd variations, yielding more reliable crowd density maps. %for crowd counting.

%namely Crowd Scale Equivariance (CSE) and Crowd Entropy Consistency (CEC).
%we proposed two self-supervised 

%{\color{red}@Haoran, can you carry on? 2-3 sentences. please highlight the Unsupervised pathway, since the next subtitle is it. 

%\begin{enumerate}
%\item After supervised initialisation,
%\item pseudo labels were generated with help of proposed GCRU, which explored the invariant representation from crowd images suffering different variations.
%\item we proposed two self-supervised regularization based on the crowd consistency in substantial unlabeled data to further reduce the effect of crowd variations for model training.
%\end{enumerate}
%}

In the semi-supervised settings, the training dataset %in each crowd dataset is 
was split into two subsets, namely the labeled set $ \mathcal{X}^{{l}} = \{\mathbf{x}^{l}_{n},\mathbf{y}^{l}_{n}\}_{n=1}^{N^{l}}$  and the unlabeled set $\mathcal{X}^{{u}} = \{\mathbf{x}^{u}_{m}\}^{N^{u}}_{m=1}$, where $N^{l}$ is sample number of labeled data and $N^{u}$ is sample number of unlabeled data. 
%The crowd variations from real world contains head scales, illumination, background, etc.
}

\subsection{Unsupervised Pathway}

In Fig.\ref{fig:unsup-path}, we demonstrate the general idea of unsupervised pathway in our semi-supervised framework. 
Here we aim at generating pseudo labels for the unlabeled set $\mathcal{X}^{{u}}$ such that the trained representation can be less sensitive to some variations. 
%To achieve this, we proposed two self-supervised regularization terms (Crowd Scale Equivariance and Crowd Entropy Consistency) on modelling the variations, and the 
%our goal is to take advantages of unlabeled crowd data $\mathcal{X}^{\math{u}}$. 
Two self-supervised regularization terms (i.e., CSE (Eq.4) and CEC(Eq.7)) were proposed, based on which an augmentation sequence can be generated to simulate the crowd variations, before applying GCRU for pseudo label generation.

%derived to reduce the effect of crowd variations, 
%{\color{red}@Haoran, i felt you should add one line for the K+2 sequence}{\color{blue} and a sequence of augmentation was applied to simulate the different variations. Then, we proposed a Gated-Crowd-Recurrent-Unit to learn the sequence-wise crowd representation.} 
%. %We demonstrated the details below.

\paragraph{Crowd Scale Equivariance (CSE)} 
Modelling variations such as head scale \cite{hu2020count,luo2020hybrid} can be an effective way for improved crowd density estimation, yet both works relied heavily on labelled data. 
%Most recent works \cite{hu2020count,jiang2020attention} trained the deep CNN on handling the head scale variation with improved performance for crowd density estimation, yet their methods required substantial labeled data. 
%{\color{red} @Haoran, can you confirm both works didn't use self-supervised or unsupervised learning on modeling head variation? }
Motivated by this,
%In contrast, 
we also modelled the head scale variation, but in an unsupervised manner.
The lack of labels makes it a challenging modelling task and here we proposed a loss (or regularisation term), named 
Crowd Scale Equivariance (CSE), based on self-supervised learning concept. 
The rationale behind is that the crowd density distribution (and the counting result) should not be affected by different head/image scales. 
%the scale of crowd images shouldn't affect the distribution of crowd densities.
%In our approach, the head scale variation was considered in a Main Variation Block (MVB in Fig.2) without any annotation of data. Missing the labels makes it hard for model to recognize the individual scale of each head. Here we proposed a regularization of Crowd Scale Equivariance (CSE) with leveraging only the unlabeled data $\mathcal{X}^{\math{u}}$.
%Intuitively, an independent crowd image in different sizes (sizes of heads also varied) represent the same density distribution and crowd counting, which introduces the crowd scale equivariance. 
%The regularization of CSE leads model to be more robust and less sensitive of head scale variation. 
Given unlabeled crowd image sample $\mathbf{x}_m^{{u}}$, and a re-scaling transformation function $\mathscr{R}$ (with re-scaling rate $r$), the CSE loss can be defined as
\begin{equation}
\mathcal{L}_{CSE}= \| \mathscr{D}(\mathscr{F}(\mathscr{R}(\mathbf{x}^{{u}}_{m})))-\mathscr{R}(\mathscr{D}(\mathscr{F}(\mathbf{x}^{u}_{m}))) \|_{1}
\label{eq:L_cse}
\end{equation}
%to be minimised.
%{\color{red} @Haoran, double check, why the second $\mathscr{R}$ is not outside $\mathscr{D}$? doesn't align well to the rationale}
It is worth noting that both 
$\mathscr{F}(\cdot)$ and $\mathscr{D}(\cdot)$ do not include dense layers %include only convolution operations, which makes it 
and they are flexible for different input image sizes.  
%$\mathscr{D}(\mathscr{F}(\cdot))$ can accept different sizes of images. 
We used $L_{1}$-norm because it is less sensitive to outliers \cite{ng2004feature}.
This CSE regularisation is demonstrated in the Main Variation Block (MVB) in Fig. \ref{fig:unsup-path}, which also outputs a sequence of feature maps $\mathbf{G}_{MVB}^{\mathbf{x}^u_{m}}$ for future processing: 
\begin{equation}
\begin{aligned}
\mathbf{G}_{MVB}^{\mathbf{x}^u_{m}} = \{ & \mathscr{U}(\mathscr{F}(\mathscr{R}(\pmb{x}^{{u}}_{m}))), \mathscr{F}(\pmb{x}^{{u}}_{m})\}, \\
\end{aligned}
\end{equation}
where $\mathscr{U}(\cdot)$ is an up-sampling function. 

\paragraph{Crowd Entropy Consistency (CEC)}
Consistency regularisation is a popular notion in semi-supervised learning for robust model training \cite{verma2019interpolation,li2020dividemix}, and the rationale behind is that the target image's semantics should keep unchanged during image transformations. 
It is worth noting that CSE is a special case when the transformation is image scaling.

% {\color{blue}{only scale? CEC used K+2 augmentations. This sentence may not be necessary}}
%Regularization of consistency is the most popular technique for robust model learning in state-of-the-art semi-supervised works \cite{verma2019interpolation,li2020dividemix}, where the rationale behind is that the target class-semantic is invariant during the image transformation.

For robust crowd modelling, we further extended this semantics-invariant concept to other transformations to reduce the effect of crowd variations such as illumination, image quality. 
As shown in Fig. \ref{fig:unsup-path}, the Auxiliary Variation Block (AVB) includes $K$ weakly augmented crowd images via image operations $\{\mathscr{A}_k(\cdot)\}^{K}_{k=1}$ such as illumination adjustment, grayscale conversion, gamma adjustment, etc.
%Similarly, 
For crowd image $\pmb{x}^{{u}}_{m}$, based on $\{\mathscr{A}_k(\cdot)\}^{K}_{k=1}$ and $\mathscr{F}(\cdot)$, AVB outputs a sequence of feature maps $\mathbf{G}_{AVB}^{\mathbf{x}^u_{m}}$: 
\begin{equation}
%\begin{aligned}
\mathbf{G}_{AVB}^{\mathbf{x}^u_{m}} = \{ 
\mathscr{F}(\mathscr{A}_1(\pmb{x}^{{u}}_{m})), \mathscr{F}(\mathscr{A}_2(\pmb{x}^{{u}}_{m})),.., \mathscr{F}(\mathscr{A}_K(\pmb{x}^{{u}}_{m}))\}.
%\label{eq:G_avb}
\end{equation}

Given the output sequences $\mathbf{G}_{MVB}^{\mathbf{x}^u_{m}}$ and $\mathbf{G}_{AVB}^{\mathbf{x}^u_{m}}$,
we further calculated the sequence expectations $\bar{\mathbf{G}}_{MVB}$ and $\bar{\mathbf{G}}_{AVB}$. 
Motivated by \cite{pan2020unsupervised}, which used entropy maps to address domain gap in unsupervised semantic segmentation tasks, 
% https://arxiv.org/abs/1904.01886
% https://openaccess.thecvf.com/content_CVPR_2019/papers/Vu_ADVENT_Adversarial_Entropy_Minimization_for_Domain_Adaptation_in_Semantic_Segmentation_CVPR_2019_paper.pdf
% https://arxiv.org/abs/2004.07703
we proposed the Crowd Entropy Consistency (CEC) regularisation/loss:
\begin{equation}
\mathcal{L}_{CEC}= \| \mathbf{E}_{MVB} - \mathbf{E}_{AVB} \|_{1}, 
\label{eq:L_cec}
\end{equation}
where $\mathbf{E}_{MVB}$ and $\mathbf{E}_{AVB}$ are the pseudo entropy maps defined as follows:
\begin{equation}
\begin{aligned}
& \mathbf{E}_{MVB}=\frac{-1}{\log C} \sum_{c \in C} \mathbf{P}_{MVB}^c \log \mathbf{P}_{MVB}^c, \\
& \mathbf{E}_{AVB}=\frac{-1}{\log C} \sum_{c \in C} \mathbf{P}_{AVB}^c \log \mathbf{P}_{AVB}^c.
\end{aligned}
\label{eq:entropy_maps}
\end{equation}
In Eq (\ref{eq:entropy_maps}), $\mathbf{P}_{MVB}^c = \sigma(\bar{\mathbf{G}}_{MVB}^{c})$ and 
$\mathbf{P}_{AVB}^c = \sigma(\bar{\mathbf{G}}_{AVB}^{c})$ where $\sigma(\cdot)$ is the sigmoid function; 
$C$ is the feature map number, and $\log C$ is a normalisation term.

The motivation of $\mathcal{L}_{CEC}$ is from the information gain concept in information theory. 
For decision tree, one of the tree-splitting principle is to select the most discriminant feature with the maximal information gain (measured in terms of entropy differences) \cite{kingsford2008decision}. However, in our case, there should not be any significant entropy differences (i.e., information gain) between two different types of transformed images (via $\mathscr{R}(\cdot)$ and $\{\mathscr{A}_k(\cdot)\}^{K}_{k=1}$) in the "variation-insensitive" feature space. 
In this case, $\mathcal{L}_{CEC}$ can be used as a regularisation term to minimise the impact of crowd variations. 

\begin{figure}
\centering
\includegraphics[width=\linewidth]{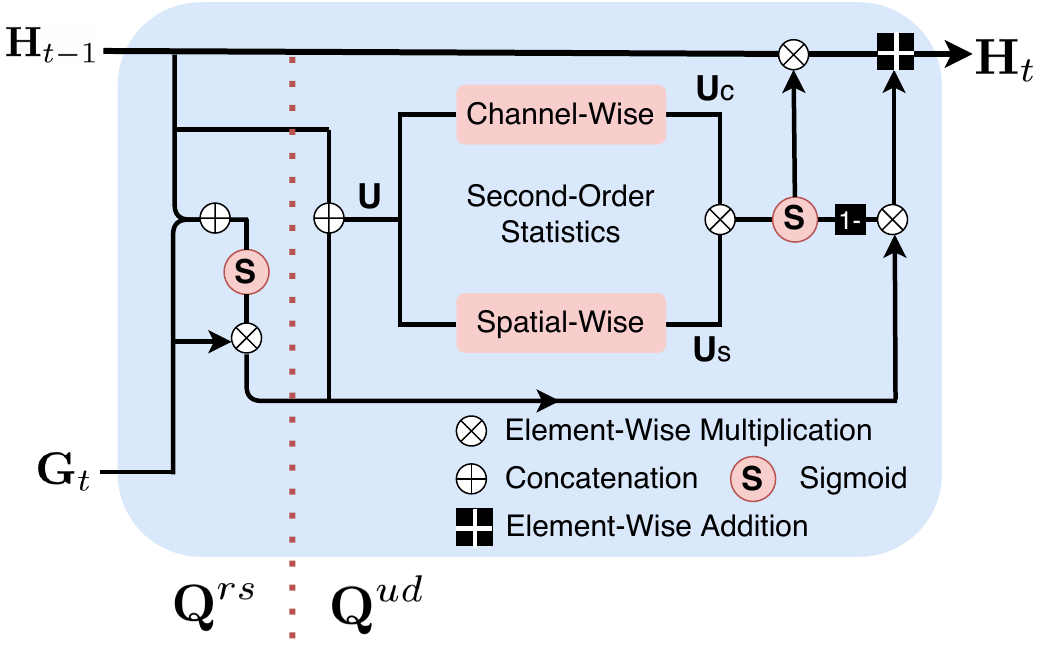}
\caption{Gated-Crowd-Recurrent-Unit}
\label{fig:GCRU}
\end{figure}

\paragraph{Gated-Crowd-Recurrent-Unit} 
For an unlabeled image $\mathbf{x}^u_{m}$, we generated the pseudo label for better training effect.
Given sequence $\mathbf{G}(\mathbf{x}^u_{m}) = \{\mathbf{G}_{MVB}^{\mathbf{x}^u_{m}}, \mathbf{G}_{AVB}^{\mathbf{x}^u_{m}} \}$, consisting of the outputs from MVB and AVB in Fig. \ref{fig:unsup-path}, we propose Gated-Crowd-Recurrent-Unit (GCRU) for sequence encoding. The internal structure of GCRU can be found at Fig. \ref{fig:GCRU}. Compared with standard deep model for sequential data (e.g., GRU \cite{cho2014learning}), our GCRU not only can preserve the high-order tensor structure, but also can extract the crowd-driven features for modelling. 
For simplicity, for the rest of this subsection we used $\mathbf{G}$ instead of $\mathbf{G}(\mathbf{x}^u_{m})$.

%$\mathbf{G}_{MVB}^{\mathbf{x}^u_{m}}$ and $\mathbf{G}_{AVB}^{\mathbf{x}^u_{m}}$ from MVB and AVB in Fig.\ref{fig:unsup-path}, we proposed a crowd-knowledge-driven recurrent unit (Fig. \ref{fig:GCRU}) named Gated-Crowd-Recurrent-Unit (GCRU) to encode %$\mathbf{G}(\mathbf{x}^u_{m})$ defined as:
%As the Pseudo-Labeling-based framework, high-quality pseudo labels were generated for unlabeled data, which should be invariant (counting and density) to different crowd variations. With the outputfrom MVB and AVB 
%, we can obtain the sequence of crowd representation, and the sequence of representation are associated closely, which represent the same crowd distribution and same number of pedestrians, which formed as:
%\begin{equation}
%\mathbf{G}(\mathbf{x}^u_{m}) = \{\mathbf{G}_{MVB}^{\mathbf{x}^u_{m}}, \mathbf{G}_{AVB}^{\mathbf{x}^u_{m}} \},
%\end{equation}
%which has a size of $K+2$ ($2$ from MVB and $K$ from AVB).
%From simplicity, we used $\mathbf{G}$ instead. 

GRU \cite{cho2014learning} is a popular sequential modelling method, yet it takes vector sequence as input while in our case, the input is high-order tensor sequence $\mathbf{G}=\{\mathbf{G}_t \in \mathbb{R}^{C \times H \times W}\}_{t=1}^{K+2}$ . 
Our GCRU can preserve the tensor structure of the crowd data, and given current input $\mathbf{G}_{t}$ and previous state $\mathbf{H}_{t-1}$ it can calculate the current high-order hidden state $\mathbf{H}_t \in \mathbb{R}^{C \times H \times W}$ 
\begin{equation}
\mathbf{H}_{t} := \text{GCRU}(\mathbf{G}_{t}, \mathbf{H}_{t-1}), \ t \in \{1,2..,K+2\}.
\end{equation}

Similar to GRU, our GCRU also has two gates, namely the reset gate ($\mathbf{Q}^{rs}\in \mathbb{R}^{C \times H \times W}$) and the update gate ($\mathbf{Q}^{ud}\in \mathbb{R}^{C \times H \times W}$). 
The reset gate is used to reduce noises in $\mathbf{G}_{t}$, and refined signal $\mathbf{\hat{G}}_{t} \in \mathbb{R}^{C \times H \times W}$ can be calculated via: 
\begin{equation}
\mathbf{\hat{G}}_{t} :=\mathbf{Q}^{rs} \odot \mathbf{G}_{t},\text{where} \ \mathbf{Q}^{rs} :=\sigma(\text{Conv}([\mathbf{G}_{t} \oplus \mathbf{H}_{t-1}])), \\
\end{equation}
where $\odot$ is the entry-wise multiplication; $\oplus$ is concatenation; $\text{Conv}(\cdot)$ is a convolutional operation for dimensionality reduction and $\sigma(\cdot)$ denotes the sigmoid function.  %and $\oplus$ is concatenation. 
Then the update gate can be applied between $\mathbf{\hat{G}}_t$ and $\mathbf{H}_{t-1}$.

In the most recent work \cite{duan2020sofa}, the authors demonstrated the effectiveness of second-order statistics in crowd modelling. They calculated the channel-wise covariance matrix, which was converted into correlation maps on learning the crowd representation.
% In the most recent work \cite{duan2020sofa}, the authors demonstrated the effectiveness of second-order statistics in crowd modelling, and they calculated the channel-wise covariance matrix, before converting to correlation map for feature extraction. {\color{red}@haoran, is this right?}
% {\color{blue} In the most recent work \cite{duan2020sofa}, the authors demonstrated the effectiveness of second-order statistics in crowd modelling. They calculated the channel-wise covariance matrix, which was converted into correlation maps, on learning the crowd representation.}
Motivated by this, we proposed the crowd-driven update gate by employing the crowd second-order statistics.
In our case, based on a two-stream structure, given $\mathbf{U} =\mathbf{\hat{G}}_{t} \oplus \mathbf{H}_{t-1}$ we extracted both the channel-wise correlation maps $\mathbf{U}_{c} \in \mathbb{R}^{C \times 1 \times 1}$ \cite{duan2020sofa} and spatial-wise correlation maps $\mathbf{U}_{s} \in \mathbb{R}^{1 \times Z \times Z}$ respectively, where $Z = WH$.
Given that, the current hidden state $\mathbf{H}_{t}$ can be calculated as follows:
\begin{equation}
\begin{aligned}
& \mathbf{H}_{t} := \mathbf{Q}^{ud} \odot \mathbf{H}_{t-1} + (1-\mathbf{Q}^{ud}) \odot \mathbf{\hat{G}}_{t}, \\
& \text{where} \ 
\mathbf{Q}^{ud} := \sigma(\text{Conv}(\mathbf{U}_{c} * \mathbf{U}_{s})),\\
%& \text{where} \ 
%\mathbf{g}^{ud}_{GRU} := \sigma(W \cdot\mathbf{U})
\end{aligned}
\end{equation}
where $*$ is the tensor-multiplication. 

\textit{Pseudo Label Generation}: GCRU's output at final step
% $K+2$ (i.e., $\text{GCRU}()_{K+2}$ or $\mathbf{H}_{K+2}$) 
can be used for pseudo label generation. 
For unlabelled image $\mathbf{x}^u_{m}$, the pseudo label $\mathbf{y}^u_{m}$ can be written as:  
\begin{equation}
\mathbf{y}^u_{m} =  \mathscr{D}(\text{GCRU}(\mathbf{G}(\mathbf{x}^u_{m})_{K+2})),
\end{equation}
based on which the pseudo supervised (PS) loss $\mathcal{L}_{PS}$ can be calculated via:

%The final sequence-wise representation $\mathbf{J}_{K+2}$ is used to generate ($\mathscr{D}(\mathbf{J}_{K+2})$) the pseudo ground truth density map $\{\mathbf{y}^{u}_{m}\}^{N^{u}}_{m=1}$ for $\mathcal{X}^{u}$, and we also built a Pseudo Supervised (PS) $L_{2}$ loss to supervise the model, which is calculated as:

\begin{equation}
\mathcal{L}_{PS} = \frac{1}{WH} \sum_{w=1}^{W} \sum_{h=1}^{H}\left(\mathscr{D}(\mathscr{F}(\mathbf{x}^{{u}}_{m}))_{w,h} - \mathbf{y}^{{u}}_{m,w,h}\right)^{2}.
\label{eq:L_ps}
\end{equation}

\subsection{Training Strategy} 
With supervised loss $\mathcal{L}_{S}$ (as defined in Eq. (\ref{eq:L_s})) and unsupervised losses $\mathcal{L}_{CSE}, \mathcal{L}_{CEC}, \mathcal{L}_{PS}$ (as defined in Eq. (\ref{eq:L_cse}), Eq. (\ref{eq:L_cec}), Eq. (\ref{eq:L_ps}) respectively), we can perform semi-supervised learning.
Given labeled set  $\mathcal{X}^{l}= \{\mathbf{x}^{l}_{n},\mathbf{y}^{l}_{n}\}_{n=1}^{N^{l}}$ and unlabeled set $\mathcal{X}^{u}= \{\mathbf{x}^{u}_{m}\}^{N^{u}}_{m=1}$,
we constructed the following joint loss function to be minimised:
\begin{equation}
\begin{aligned}
&\mathcal{L} = \mathbb{E}_{\mathbf{x}^{l}_{n},\mathbf{y}^{l}_{n} \sim \mathcal{X}^{l}}\mathcal{L}_{S}(\mathbf{x}^{l}_{n},\mathbf{y}^{l}_{n}) +\\
&\lambda \mathbb{E}_{\mathbf{x}^{u}_{m},\mathbf{y}^{u}_{m} \sim \mathcal{X}^{u}}  \{\mathcal{L}_{CSE}(\mathbf{x}^{u}_{m}) + \mathcal{L}_{CEC}(\mathbf{x}^{u}_{m}) + \mathcal{L}_{PS}(\mathbf{x}^{u}_{m},\mathbf{y}^{u}_{m})\},
\end{aligned}
\label{eq:L_joint}
\end{equation}
where $\lambda$ plays the role in balancing the supervised loss and unsupervised losses. 
In Eq. (\ref{eq:L_joint}), it is obvious that supervised training becomes dominant with smaller values of $\lambda$, and vice versa.
Model performance with respect to $\lambda$ were studied in the experiments section.
%For example, when $\lambda=0$ our method becomes supervised method when 

In our semi-supervised learning framework, the unsupervised losses were constructed based on self-supervised schemes to minimise the impact of crowd variations, while a supervised loss was designed to learn better crowd information.
These proportion of different types of losses should be dynamic at different training stages. 
Given that, we employed the dynamic $\lambda$ training strategy, as shown in Algorithm \ref{algo:lambda}.  

\begin{algorithm}
\SetAlgoLined
\KwIn{ 
\begin{itemize}
\item[-] Labeled set $\mathcal{X}^{l}= \{\mathbf{x}^{l}_{n},\mathbf{y}^{l}_{n}\}_{n=1}^{N^{l}}$;
\item[-] Unlabeled set $\mathcal{X}^{u}= \{\mathbf{x}^{u}_{m}\}^{N^{u}}_{m=1}$;
\end{itemize}
}
\KwOut{Model parameters $\pmb{\Theta}$ (corresponding to $\mathscr{F}(\cdot)$, $\mathscr{D}(\cdot)$, and $\text{GCRU}(\cdot)$)}
Initialisation\;
 \For{$b = 1 \ to \ B$}{
 Generating pseudo labels: $\{\mathbf{y}^{u}_{m}\}^{N^{u}}_{m=1}$; \\
  $\lambda = \frac{b - 1}{B}$; \\
  $\mathcal{L}^{b} = \mathbb{E}_{\mathbf{x}^{l}_{n},\mathbf{y}^{l}_{n} \sim \mathcal{X}^{l}}\mathcal{L}_{S}(\mathbf{x}^{l}_{n},\mathbf{y}^{l}_{n}) +  $
  $\lambda \mathbb{E}_{\mathbf{x}^{u}_{m},\mathbf{y}^{u}_{m} \sim \mathcal{X}^{u}} \{\mathcal{L}_{CSE}(\mathbf{x}^{u}_{m}) + \mathcal{L}_{CEC}(\mathbf{x}^{u}_{m}) + \mathcal{L}_{PS}(\mathbf{x}^{u}_{m},\mathbf{y}^{u}_{m})\}$; \\
  Calculating gradients: $\triangledown_{\pmb{\Theta}}^b \leftarrow \frac{\partial \mathcal{L}^{b}}{\partial \pmb{\Theta}}$;\\ 
  Updating model: $\pmb{\Theta}\leftarrow (\pmb{\Theta}, \triangledown_{\pmb{\Theta}}^b)$
}
 \caption{Training Strategy (Dynamic $\lambda$)}
 \label{algo:lambda}
\end{algorithm}

In Algorithm \ref{algo:lambda}, the joint loss $\mathcal{L}^{b}$ changes w.r.t training epoch.
Specifically, in the first epoch (i.e., $b=1$), due to the randomly generated pseudo labels, the coefficient of the unsupervised loss is set to zero (with $\lambda=0$). 
With the increasing training epoches, the unsupervised loss would become more important (which can reduce the effect of crowd variants), yielding more reliable pseudo labels (for unsupervised learning in further epoches).
With a large total epoch, unsupervised loss would nearly have the same weight as the supervised loss, serving as a domain-knowledge driven regularisation term for robust crowd density estimation. 

\section{Experiments}

\begin{table}[htp]
\centering
\caption{Semi-supervised algorithm comparison under the settings of\cite{sindagi2020learning}}
\scalebox{0.7}{
\begin{tabular}{ |c|c||c|c|c|}
 \hline
  \qquad \quad \quad \ \ \ \ \ - & \ \quad \ \ Split(\%)& UCF\_QNRF & ShanghaiTech A& ShanghaiTech B \\
 \hline
 \qquad \quad \ \ \ Method & \quad $N^{l}(\%)$, \ $N^{u}(\%)$ & \ MAE$\downarrow$\quad MSE$\downarrow$ & \ MAE$\downarrow$\quad MSE$\downarrow$ & \ MAE$\downarrow$\quad MSE$\downarrow$ \\
 \hline\hline

 GP \cite{sindagi2020learning}& \quad \quad 5\%, \ \ \ 95\% &  \ 160.0\quad \ \  275.0  &  \ 102.0\quad \ \  172.0  &  \ 15.7\quad \quad 27.9   \\
 
 \textbf{Ours (S4-Crowd)}& \quad  \quad 5\%, \ \ \ 95\% &  \ \textbf{158.6}\quad \ \  \textbf{269.6}  &  \ \textbf{100.8}\quad \ \  \textbf{170.3}  &  \ \textbf{15.3}\quad \quad \textbf{24.4}  \\\cdashline{1-5}[2pt/2pt]
 
 GP \cite{sindagi2020learning}& \quad  \quad 25\%, \ 75\% &  \ 147.0\quad \ \  226.0  &  \ 91.0\quad \quad 149.0  &  \quad - \qquad \quad - \\
 
 \textbf{Ours (S4-Crowd)}& \quad  \quad 25\%, \ 75\% &  \ \textbf{140.5}\quad \ \  \textbf{224.8}  &  \ \textbf{81.4}\quad \quad \textbf{137.9}  &  \ \textbf{12.9}\quad \quad \textbf{21.6}  \\
 \cdashline{1-5}[2pt/2pt]
 
 GP \cite{sindagi2020learning}& \quad  \quad 50\%, \ 50\% &  \ 136.0\quad \ \  218.0  &  \ 89.0\quad \quad 148.0  &  \quad - \qquad \quad - \\

 \textbf{Ours (S4-Crowd)}& \quad  \quad 50\%, \ 50\% &  \ \textbf{119.9}\quad \ \  \textbf{214.3}  &  \ \textbf{74.1}\quad \quad \textbf{125.3}  &  \ \ \textbf{9.8}\quad \quad \  \textbf{19.2}  \\
 \cdashline{1-5}[2pt/2pt]
 
 GP \cite{sindagi2020learning}& \quad  \quad 75\%, \ 25\% &  \ 129.0\quad \ \  210.0  &  \ 88.0\quad \quad 139.0  &  \quad - \qquad \quad - \\
 
 \textbf{Ours (S4-Crowd)}& \quad  \quad 75\%, \ 25\% &  \ \textbf{114.6}\quad \ \  \textbf{198.7}  &  \ \textbf{68.3}\quad \quad \textbf{117.3}  &  \ \ \textbf{9.7}\quad \quad \  \textbf{17.8}   \\
 \hline
 \end{tabular}}
 
 \label{tab:comp-GP}
\end{table}
 
 \begin{table}[htp]
\centering
 \caption{Semi-supervised algorithm comparison under the settings of \cite{liu2020semi} }
\scalebox{0.64}{
\begin{tabular}{ |c||c|c|c|c|c|}
 \hline
  \qquad \qquad \qquad \qquad \quad \ \ - & \quad \ \ UCF\_QNRF & \ \  ShanghaiTech A& \ \ \ ShanghaiTech B& \quad WE \\
 \hline
 \qquad \qquad \qquad \quad \quad  \ \ Split & \ $N^{l}(721)$, $N^{u}(480)$ & \ \ $N^{l}(90)$, $N^{u}(210)$ & \ $N^{l}(120)$, $N^{u}(280)$ & \ $N^{l}(947)$, $N^{u}(2433)$  \\
 \hline
 
 \qquad \qquad \qquad \qquad  Method & \ \quad MAE$\downarrow$\quad MSE$\downarrow$ & \ \quad MAE$\downarrow$\quad MSE$\downarrow$ & \ \quad MAE$\downarrow$\quad MSE$\downarrow$& \ \quad MAE$\downarrow$\quad MSE$\downarrow$  \\
 \hline\hline

 L2R \cite{liu2018leveraging}&  \quad \ 148.9\quad \ \  249.8  & \quad \ 90.3 \ \quad \ \  153.5  &  \quad \ 15.6\quad \quad  24.4&  \quad \ 13.9\quad \quad \ \ - \\
 
 UDA \cite{xie2019unsupervised}&  \quad \ 144.7\quad \ \  255.9  & \quad \  93.8 \ \quad \ \  157.2  &  \quad \ 15.7\quad \quad  24.1&  \quad \ 14.2\quad \quad \ \ - \\
 
 MT \cite{tarvainen2017mean}&  \quad \ 145.5\quad \ \  250.3  &  \quad \ 94.5 \quad \ \ \ 156.1  &  \quad \ 15.6\quad \quad  24.5&  \quad \ 14.1\quad \quad \ \ - \\
 
 ICT \cite{verma2019interpolation}&  \quad \ 144.9\quad \ \  250.0  & \quad \  92.5 \ \quad \ \  156.8  &  \quad \ 15.4\quad \quad  23.8&  \quad \ 14.9\quad \quad \ \ - \\
 \cdashline{1-5}[2pt/2pt]
 
 IRAST \cite{liu2020semi} &  \quad \ 135.6\quad \ \  233.4  &  \quad \ 86.9\quad \quad  148.9  &  \quad \ 14.7\quad \quad  22.9&  \quad \ 11.1\quad \quad \ \ - \\
 
 IRAST(SPN) \cite{liu2020semi} &  \quad \ 128.4\quad \ \  225.3  &  \quad \ 83.9\quad \quad 140.1  &  \quad \ \quad - \qquad \quad -&  \quad \quad -\quad \quad \quad \ - \\
 
 \textbf{Ours (S4-Crowd)} &  \quad \ \textbf{127.0}\quad \ \ \textbf{211.1}  &  \quad \ \textbf{77.8}\quad \quad  \textbf{128.8}  &  \quad \ \textbf{10.9}\quad \quad \textbf{17.3}  &  \quad \ \ \textbf{9.3}\quad \ \quad \ \ - \\
 \hline
 \end{tabular}
 }
 \label{tab:comp-IRAST}
 \end{table}

\subsection{Datasets}

We evaluated our approach on four popular crowd counting datasets, namely, \textbf{ShanghaiTech A/B} \cite{Zhang_2016_CVPR}, \textbf{UCF\_QNRF} \cite{idrees2018composition}, and  \textbf{World Expo’10 (WE)} \cite{zhang2015cross}. 
%with comparing to the recent state-of-the-art works. 
Specifically, \textbf{ShanghaiTech} \cite{Zhang_2016_CVPR} contains two parts, part A and part B. 
Part A was randomly collected from internet of congested street view with 300 images in the training set and 182 images in the test set. 
Part B was collected from similar street views with relatively sparse crowd, and it has 400 images in the training set and 316 images in the test set. 
\textbf{UCF\_QNRF} \cite{idrees2018composition} is a large-scale dataset with 1535 high-resolution images, with 1201 images in the training set and 334 images in the test set.
Both \text{ShanghaiTech} and \text{UCF\_QNRF} were collected in various crowd scenes. \textbf{World Expo’10 (WE)} dataset \cite{zhang2015cross} was collected in fixed crowd scenes with 3380 training images and 600 testing images.

For fair algorithm comparison, following other semi-supervised learning approaches in the crowd counting community \cite{liu2020semi,sindagi2020learning}, we kept the test set unchanged while splitting the training set into labeled set (with size $N^l$) and unlabeled set (with size $N^u$).  
More details of the training set splitting can be found in respective experiments below. %\ref{tab:comp-GP}-\ref{tab:comp-IRAST}.
 
%{\color{red} In our semi-supervised settings, following [ref], we xxxxxxxx. 
%Following the most semi-supervised learning \cite{berthelot2019mixmatch}, the labeled and unlabeled data was selected randomly.
%}

\subsection{Implementation Details}
VGG16 with first 13 convolutional layers was a popular backbone for CNN-based crowd counting \cite{Cheng_2019_ICCV,jiang2020attention}, which was also used in this work (i.e., $\mathscr{F}(\cdot)$). 
%was used as our model to learn the crowd representation, which was convention in many state-of-the-art works as baseline \cite{Cheng_2019_ICCV,jiang2020attention}.
For density map generator $\mathscr{D}(\cdot)$, we used 3 convolutional layers and 1 up-sampling layer. %to reduce the dimension and one up-sampling layers to control the size of predicted density map. 
%In this work, we aim to design a semi-supervised crowd counting framework, the representation learning and density regression models will be explored in the future. 
The model was optimised by Adam \cite{kingma2014adam} with batch size $30$, and learning rate $1e-5$.

For image transformations (i.e.,$\{\mathscr{A}_k(\cdot)\}^{K}_{k=1}$), we used $K=5$ image operations including grayscale conversion, gamma adjustment, illumination adjustment (bright/dark), perspective adjustment.

\subsection{Evaluation Metrics}

Mean Absolute Error (MAE) and Mean Square Error (MSE) are two most popular evaluation metrics for performance evaluation \cite{Cheng_2019_ICCV,jiang2020attention}, which also used in this work. 
Given the predicted crowd counts of $N$ testing images $\{{e}^{pr}_{i}\}_{i=1}^N$ (can be inferred from Eq. (\ref{eq:inference}) with $\mathscr{F}(\cdot)$ and $\mathscr{D}(\cdot)$) and the corresponding ground truth counts $\{{e}^{gt}_{i}\}_{i=1}^N$,
MAE/MSE can be written as follows:

\begin{footnotesize}
\begin{equation}
MAE=\frac{1}{N} \sum_{i=1}^{N}\left|{e}^{pr}_{i}-{e}^{gt}_{i}\right|, MSE=\sqrt{\frac{1}{N}\sum_{i=1}^{N}\left({e}^{pr}_{i}-{e}^{gt}_{i}\right)^{2}}.
\end{equation}
\end{footnotesize}

\section{Results and Discussions}

In this section, we qualitatively and quantitatively evaluate our proposed methods with in-depth discussions.

\subsection{Comparison with existing Methods}

\subsubsection{\textbf{\textit{{Comparison with GP}} }}
Sindagi et al. proposed a GP-based semi-supervised method \cite{sindagi2020learning} and evaluated it in a few different settings (i.e., in terms of the labeled-set/unlabeled-set split in $\%$ of the training set). 
Following their settings \cite{sindagi2020learning}, on three datasets UCF\_QNRF, ShanghaiTech A, and ShanghaiTech B we compared GP-based approach with our $S^{4}Crowd$ method. 
As shown in Table \ref{tab:comp-GP}, our method outperformed GP-based method in different settings on the three datasets, and the performance gain tended to be larger on UCF\_QNRF and ShanghaiTech A.
For ShanghaiTech B, since it is a simple dataset with relatively sparse crowd, the lower error limit can be easily hit. Nevertheless our method still obtained higher results. It is worth to note that with an extremely limited amount of labeled data (i.e., $5\%$), it is challenging to train a model as good as in other settings. In order to be fair in real-world scenarios, we randomly select the labeled data in this work and the selection mechanism will be considered in the future, which may be useful for crowd counting in extreme situations. \\

\subsubsection{\textbf{\textit{{Comparison with IRAST}}}} Recently, Liu et al. proposed a semi-supervised crowd counting by leveraging surrogate tasks \cite{liu2020semi}, and evaluated it on a number of datasets.
In their settings, they randomly chose labeled set with size $N^l=90$ in ShanghaiTech A; $N^l=120$ in ShanghaiTech B; $N^l=721$ in UCF\_QNRF; and $N^l=947$ in WE, and used corresponding dataset's rest training set as unlabeled set.
These settings as well as a number of semi-supervised learning algorithms' performance can be found in Table \ref{tab:comp-IRAST}.

Out of the algorithms,
Unsupervised Data Augmentation (UDA) \cite{xie2019unsupervised}, 
Mean teacher (MT) \cite{tarvainen2017mean} 
and  Interpolation Consistency Training (ICT) \cite{verma2019interpolation} were three highly popular semi-supervised methods.
For crowd counting tasks, Liu et al. modified them by using the density map as output \cite{liu2020semi}.
Learning to Rank (L2R) \cite{liu2018leveraging} was a self-supervised crowd modelling method that can leverage unlabeled crowd images from Internet, and Liu et al. \cite{liu2020semi} also changed it to the same semi-supervised setting as shown in Table \ref{tab:comp-IRAST}. 

We compared these algorithms as well as IRAST, IRAST(SPN) \cite{liu2020semi} with our $S^{4}Crowd$ method, and ours outperformed all the algorithms. 
Compared with the most recent IRAST model \cite{liu2020semi}, our method had a $8.6$ MAE reduction on UCF\_QNRF; 9.1 MAE reduction on ShanghaiTech A; 3.8 MAE reduction on ShanghaiTech B and 1.8 MAE reduction on WE. 
It is worth mentioning that Liu et al. also combined their IRAST with state-of-the-art supervised model SPN\cite{chen2019scale} for further error reduction. 
In this case, even without external supervised model, our $S^{4}Crowd$ approach still performed better.

% \subsubsection{\textbf{\textit{Feasibility of pseudo density}}} We also conducted a experiment to evaluate the feasibility of using the pseudo ground truth density for unlabeled data. As it is shown in Fig. \ref{fig:fopd}, based on ShanghaiTech part A, we visualized the normalized errors of pseudo ground truth density maps and directly predicted density maps, which were compared to the real ground truth. The pseudo density acts as the more accurate supervision of unlabeled data, and using it to train the model can improve the generalibility when labeled data is limited.

% \begin{figure}
% \centering
% \includegraphics[width=0.6\linewidth]{attach/fopd.pdf}
% \caption{Blue: Errors of VGG16 baseline prediction. Orange: Errors of our pseudo ground truth.}
% \label{fig:fopd}
% \end{figure}

\subsection{Ablation Study}
ShanghaiTech A was a popular dataset for ablation studies in the crowd counting research community \cite{bai2020adaptive,jiang2020attention}, and it was also used in our study.
We empirically set the size (in $\%$) of labeled data (resp. unlabeled data) to ${N}^{l}=15\%$ (resp. ${N}^{u}=85\%$). \\

\begin{table}[h!]
\centering
 \caption{The effect of $r$ for scaling operation $\mathscr{R(\cdot)}$.}
\begin{tabular}{ |c||c||c|c|  }
 \hline
 \multicolumn{4}{|c|}{ShanghaiTech A} \\
 \hline
 \quad \ \ $\mathscr{R(\cdot)}$ &$N^{l}(\%)$, \ $N^{u}(\%)$ & MAE &MSE\\
 \hline
 \quad $r = 0.1$ & \quad \ $15\%$, \ $85\%$	& 92.3 	& 159.2 \\
 \quad $r = 0.2$ & \quad \ $15\%$, \ $85\%$	& 89.6	& 159.3\\
 \quad $r = 0.4$ & \quad \ $15\%$, \ $85\%$	& 86.5	& 155.4\\
 \quad $r = 0.6$ & \quad \ $15\%$, \ $85\%$	& \textbf{82.1}	&\textbf{143.7} \\
 \quad $r = 0.8$ & \quad \ $15\%$, \ $85\%$	& 88.7	& 160.2\\
  \hline
 \end{tabular}

 \label{tab:ablation-r}
 \end{table}

\subsubsection{\textbf{\textit{Effect of re-scaling rate $r$}}} 

The scale variation was modelled with a CSE loss/regularisation in the unsupervised pathway, which was controlled by a re-scaling factor $r$.
Here we focused on the challenging down-sampling scenarios (with low resolution crowd heads/images) by setting the re-scaling factor $r$ at the range of $(0,1)$.
Specifically, we studied $r=\{0.1, 0.2, 0.4, 0.6, 0.8\}$, and the results in Table \ref{tab:ablation-r} suggested that with moderate value ($r = 0.6$), our $S^{4}Crowd$ method reached the best results. \\

\subsubsection{\textbf{\textit{Effect of $K$ augmentations}}} Our $S^{4}Crowd$ method also modelled different crowd variations by employing $K$ image operations (aka augmentations). 
Table \ref{tab:ablation-K} showed the performance with different $K$ image operations. 
We empirically found that $K=3$ (by setting to "grayscale, bright, dark") yielded the lowest MAE. \\

\begin{table}[!h]
\centering
\caption{The effect of $K$ image operations on modelling crowd variations.
 $K$=1 (grayscale), $K$=2 (grayscale, bright), $K$=3 (grayscale, bright, dark), $K$=4 (grayscale, bright, dark, gamma adjustment), $K$=5 (grayscale, bright, dark, gamma adjustment, perspective adjustment)}
\begin{tabular}{ |c||c||c|c|  }
 \hline
 \multicolumn{4}{|c|}{ShanghaiTech A} \\
 \hline
 \quad Method &$N^{l}(\%)$, \ $N^{u}(\%)$ & MAE &MSE\\
 \hline
 \quad $K = 1$ &  \quad \ $15\%$, \ $85\%$	& 86.1	& 142.9 \\
 \quad $K = 2$ &  \quad \ $15\%$, \ $85\%$	& 85.4	& \textbf{142.1} \\
 \quad $K = 3$ &  \quad \ $15\%$, \ $85\%$	& \textbf{82.1}	& 143.7\\
 \quad $K = 4$ &  \quad \ $15\%$, \ $85\%$	&  86.2	& 151.5\\
 \quad $K = 5$ &  \quad \ $15\%$, \ $85\%$	& 95.3	& 188.5 \\
  \hline
 \end{tabular}
 \label{tab:ablation-K}
 \end{table}

\subsubsection{\textbf{\textit{Effect of each component}}} 
Our $S^{4}Crowd$ framework includes many components (such as CSE loss, CEC loss, PS loss, GCRU), and it is important to study the effectiveness of them. 
In Table \ref{tab:ablation-arch}, we can observe supervised approach alone (VGG16) with only limited labels yielded the worst results.
On the other hand, VGG16's performance can be substantially improved even with the simplest self-supervised regularisation terms (V+CSE, or V+CEC), which suggested the necessity of leveraging unlabeled data (on modelling crowd variations).
Our framework's error can be further reduced with more proposed components (e.g., with PS loss, GCRU). 

To study the effect of GCRU,
we replaced GCRU by GRU, and reshaped the corresponding high-order-tensor sequences into vector sequences as input. 
The increased errors suggested the effectiveness of the crowd-driven GCRU in crowd counting tasks.
We also changed our backbone from VGG16 to advanced CSRNet \cite{li2018csrnet} and CAN \cite{liu2019context}, and the lower errors suggested the flexibility of our $S^{4}Crowd$ framework. \\

\begin{table}[!h]
\centering
 \caption{Effect of each component in our $S^{4}Crowd$}
\scalebox{0.7}{
\begin{tabular}{ |c||c||c|c|  }
 \hline
 \multicolumn{4}{|c|}{ShanghaiTech A} \\
 \hline
  Method &$N^{l}(\%)$, \ $N^{u}(\%)$ & MAE &MSE\\
 \hline
 VGG16(V) &   \ $15\%$, \ $0\%$	& 107.5	& 211.3\\
 V + CSE &   \ $15\%$, \ $85\%$	& 87.9	&155.8 \\
 V + CEC &   \ $15\%$, \ $85\%$	& 86.4	&158.1 \\
 V + GCRU + PS&  \ $15\%$, \ $85\%$  &	88.9	&164.9 \\
 V + GCRU + PS + CEC &  \ $15\%$, \ $85\%$ & 83.9	&150.3 \\
 V + GCRU + PS + CSE &   \ $15\%$, \ $85\%$ & 86.4	&149.2 \\
 V + GRU + PS + CSE + CEC &  \ $15\%$, \ $85\%$ &84.7	&147.0 \\
 V + GCRU + PS + CSE + CEC &  \ $15\%$, \ $85\%$ &82.1	&143.7 \\
 CSRNet + GCRU + PS + CSE + CEC &  \ $15\%$, \ $85\%$ &81.8	& 141.6
 \\
 CAN + GCRU + PS + CSE + CEC &  \ $15\%$, \ $85\%$ &\textbf{80.3}	& \textbf{137.4}
 \\
  \hline
 \end{tabular}
 }
 \label{tab:ablation-arch}
 \end{table}

 \subsubsection{\textbf{\textit{Effect of training strategy}}}
 We also studied the the effect of different training strategies for our semi-supervised framework $S^{4}Crowd$, and the results were reported in Table \ref{tab:ablation-lambda}. 
 We can see our dynamic $\lambda$ scheme yielded much lower errors than other strategies which used fixed values of $\lambda$. 
 
 One explanation is that the importance of unsupervised loss and supervised loss may change during different training stages. 
 At the early training stages, the generated pseudo labels were less reliable and the corresponding importance (measured by $\lambda$) should be low. 
 With more training iterations, the model benefited more from the self-supervised CSE/CEC losses (on crowd variations modelling) and supervised loss, and in this case the generated pseudo labels tended to be more reliable.
 With higher quality pseudo labels, the corresponding PS (Pseudo Supervised) loss should have a higher weight. 
 In our dynamic weight scheme, we can see the weight of the unsupervised loss (PS+CSE+CEC) increased with the more training epoches.
 At the end of the training, the unsupervised loss would have nearly the same weight as the supervised loss (see Algorithm \ref{algo:lambda}), serving as an important regularisation term in the learning tasks. \\

 \begin{table}[!h]
\centering
 \caption{Effect of the training strategy}
\scalebox{0.74}{
\begin{tabular}{ |c||c||c|c|  }
 \hline
 \multicolumn{4}{|c|}{ShanghaiTech A} \\
 \hline
 \qquad \qquad \qquad \ \ Method &$N^{l}(\%)$, \ $N^{u}(\%)$ & MAE &MSE\\
 \hline
 Joint loss in Eq.(\ref{eq:L_joint}) with $\lambda = 0$ &    \quad \ $15\%$, \ $85\%$	& 143.4	& 356.4\\
 Joint loss in Eq.(\ref{eq:L_joint}) with $\lambda = 0.1$ &    \quad \ $15\%$, \ $85\%$	& 112.8	& 240.2\\
 Joint loss in Eq.(\ref{eq:L_joint}) with $\lambda = 0.3$ &    \quad \ $15\%$, \ $85\%$	& 105.8	& 228.9 \\
 Joint loss in Eq.(\ref{eq:L_joint}) with $\lambda = 0.7$ &    \quad \ $15\%$, \ $85\%$	& 114.8	& 229.3\\
 Our dynamic $\lambda$ (Algorithm \ref{algo:lambda}) &    \quad \ $15\%$, \ $85\%$	& \textbf{82.1}	&\textbf{143.7} \\
  \hline
 \end{tabular}
 }
 \label{tab:ablation-lambda}
 \end{table}

\subsubsection{\textbf{\textit{Effect of leveraging external unlabeled data}}} 
In previous experiments, both $\mathcal{X}^{u}$ and $\mathcal{X}^{l}$ were from the same dataset.
To simulate the real-world scenarios, we also ran experiments using external unlabeled data. 
In \cite{liu2018leveraging}, the authors collected $N^{u}_{L2R}=3409$ unlabeled crowd images from Google, and used them as additional information for modelling.
For fair comparison, we employed the same network structure as their L2R method  \cite{liu2018leveraging} (i.e., same $\mathscr{F}(\cdot)$, and a different $\mathscr{D}(\cdot)$ with only 1 convolutional layer).
Since the Google crowd images used in \cite{liu2018leveraging} were not publicly available, we employed UCF\_QNRF as external unlabeled data $\mathcal{X}^{u}$.
We used ShanghaiTech A training set as $\mathcal{X}^{l}$, and used test set (i.e., with 182 images) for evaluation. 
The detailed settings and the results were reported in Table \ref{tab:comp-L2R}, from which we can see our method yielded lower MAE than L2R, even with a much smaller external $\mathcal{X}^{u}$.
We also noticed both methods did not outperform the supervised baseline substantially.
One possible reason can be the differences from two different data sources, and domain adaptation will be explored to boost the performance in the future.

\begin{table}[!hb]
\centering
 \caption{On leveraging the external unlabeled data}
\scalebox{0.7}{
\begin{tabular}{ |c||c|c||c|}
 \hline
 \qquad \ \ Method & \quad \quad \ \  $ \mathcal{X}^{l} (N^{l})$ & \quad \ \ \ $\mathcal{X}^{u}(N^{u})$ & MAE \\
 \hline
 VGG16 &  ShanghaiTech A(300) & \quad \ \ \ None(0)	& 72.8  \\
 L2R & ShanghaiTech A(300) & \ \ \ Google(3409)	& 72.0  \\
 \textbf{Ours ($S^{4}Crowd$)} & ShanghaiTech A(300) & UCF\_QNRF(1201)	& \textbf{70.3} \\
  \hline
 \end{tabular}
 }
 \label{tab:comp-L2R}
 \end{table}

\section{Conclusions}
In this paper, we proposed a semi-supervised framework $S^{4}Crowd$ for crowd counting tasks. 
Two self-supervised regularisation terms Crowd Scale Equivariance/Crowd Entropy Consistency, pseudo labeling, Gated-Crowd-Recurrent-Unit, were proposed as key components in this framework together with a dynamic training scheme for better crowd modelling. We comprehensively studied the effectiveness of each component, and our method outperformed other representative methods in various semi-supervised settings on public datasets. 
In the future, we will explore domain adaptation methods on leveraging the unlabeled crowd data in the wild.

% \section*{Acknowledgment}

% The work was supported by National Natural Science Foundation of China(Grant No.61801190), National Key Research and Development Project of China(Grant No.2019YFC0409105), Nature Science Foundation of Jilin Province(Grant No.20180101055JC), Outstanding Young Talent Foundation of Jilin Province(Grant No.20180520029JH), China Postdoctoral Science Foundation (Grant No.2017M611323), and the Fundamental Research Funds for the Central Universities, JLU.

% \section*{References}
%\bibliographystyle{unsrt} 
\bibliography{my_ref}

\end{document}